\documentclass[letterpaper, 10 pt, conference]{ieeeconf}
\usepackage{cite}
\usepackage[utf8]{inputenc}

\usepackage{hyperref}
\usepackage{caption}

\usepackage{subcaption} 
\usepackage{graphicx}
\usepackage{epstopdf}
\graphicspath{{../pdf/}{../jpeg/}{../img/}}
\DeclareGraphicsExtensions{.pdf,.jpeg,.png,.eps}

\usepackage{verbatim}
\usepackage[cmex10]{amsmath}
\usepackage{array}
\usepackage{url}

\usepackage{xcolor}%


\begin{document}
\title{\vspace*{-1cm}\LARGE \bf Towards Declarative Safety Rules for\\Perception Specification Architectures\vspace*{-5mm}}
%\title{Extending Perception Specification Architectures with Concept and Ideas of Safety}
\author{Johann Thor Mogensen Ingibergsson and Ulrik Pagh Schultz and Dirk Kraft\\
University of Southern Denmark, %The M\ae rsk Mc-Kinney M\o ller Institute \& Center for Energy Informatics\\ 
Campusvej 55, 5230 Odense M, Denmark\\
Email: \{jomo$|$ups$|$kraft\}@mmmi.sdu.dk\vspace{-3mm}}

\maketitle

\begin{abstract}
Agriculture has a high number of fatalities compared to other blue
collar fields, additionally population decreasing in rural areas
is resulting in decreased work force. These issues have
resulted in increased focus on improving efficiency of and introducing
autonomy in agriculture. Field robots are an increasingly promising
branch of robotics targeted at full automation in agriculture.
The safety aspect however is rarely addressed
in connection with safety standards, which limits the real-world applicability. In
this paper we present an analysis of a vision pipeline in connection
with functional-safety standards, in order to propose solutions for how to
ascertain that the system operates as required. Based on the analysis
we demonstrate a simple mechanism for verifying that a vision pipeline is
functioning correctly, thus improving the safety in the overall system. % \TODO{TODO}

\end{abstract}

%\begin{IEEEkeywords}
%??, ??, ??, \TODO{TODO}.
%\end{IEEEkeywords}

\section{Introduction}
\label{sec:Introduction}

Agriculture is an overexposed field in relation to injuries and
fatalities, both in the EU~\cite{patrick_j._griffin_safety_2013} and
USA~\cite{agricultural_statistics_board_agricultural_2013}. This
implies that even in developed countries there is a significant problem with
safety. In addition to the safety issues, rural areas in developing
countries are losing inhabitants to the cities, which decreases the
labour forces and puts extra emphasis on improving efficiency. This
has given birth to the idea of replacing human labour for field
work. To this end outdoor mobile robots are a solution. A subclass of
mobile robots is given by \emph{field robots}, and refers to machinery
applied for outdoor tasks, e.g., in construction, forestry and
agriculture~\cite{yang_remote_2008, ingibergsson_mmmi_2015}.
Undertaking the job of creating field robots is however a large task,
in part due to the many domains that overlap within robotics,
e.g., mechanical and software. Additionally the dynamical environment
in which a field robot operates introduces additional strain on the
robots, and as result outdoor mobile robots fail up to 10 times more
often than other types of robots~\cite{carlson_follow-up_2004}.  The
issue with field robots being more prone to failure has resulted in
research within software quality for robotics~\cite{reichardt_software_2013}. 
The same issue is present in more
mature domains, such as the avionics and automotive domains, which
adopted Model-Driven Engineering (MDE;
\cite{schmidt_guest_2006}). This adoption has led to utilizing of MDE
in robotics to improve development time and reliability, examples are
SmartSoft~\cite{steck_model-driven_2011} and for computer vision Robot
Perception Specification Language (RPSL;
\cite{hochgeschwender_declarative_2014}). These MDE approaches are however not developed
with a functional safety focus and are therefore missing some
important aspects to make the robots trustworthy in relation to
certification, and thereby improving the safety in the agricultural
industry.  For robots the vision domain is critical, this is also the
case for many other applications ranging from monitoring operation on
airfields~\cite{aguilera_visual_2006} to real-time controlling of
autonomous systems, such as vehicles~\cite{cheng_state---art_2011} and
robotics~\cite{aymeric_de_cabrol_concept_2008}. Robotics is highly
dependent on computer vision to understand and react to the
environment. This dependability imposes high constrains on the
reliability of the software and the vision pipeline. For computer
vision systems the issues are pointed out by Yang et al. ``\textit{one
  is to identify an obstacle surrounding the robot and the other it to
  determine the location of the obstacle}''~\cite{yang_human_2012}. For
field robots to assist in field work, the robots have to be safety certified
so as to minimise the liability (liability is addressed in
\cite{santosuosso_robots_2012}) of the producers.  Our goal is
to investigate if it is possible to extend an MDE approach such as RPSL to incorporate safety aspects.
Concretely we propose ideas of how to achieve sufficient safety levels
for the vision system in a field robot, and report on preliminary experiments demonstrating the viability of the
proposed methods.

%UPS: outline ikke kritisk for så kort en artikel...
%\paragraph{Outline}\TODO{TODO} This paper gives an overview of related work within vision and safety (Section \ref{sec:BackRelWork}) leading to a description of a system scenario and concept for improving safety (Section \ref{sec:ProblemDefinition}), 

\section{Background and Related Work}
\label{sec:BackRelWork}

Functional safety standards only address human dangers, e.g., ISO 26262~\cite{iso26262_road_2011} and ISO 25119~\cite{iso25119_tractors_2010}. This leaves the designer and developer
to categorise issues related to harming the robot, e.g., untraversable
ground and non-human obstacles. Hedenberg et
al.~\cite{hedenberg_safety_2011} use EN 1525 (driver-less trucks~\cite{en1525_safety_1998}), and argue that the focus on humans is
sound based on Asimov's ``\textit{A robot may not injure a human-being or,
  through inaction, allow a human-being to come to harm}''.
%\cite{asimov_i_1950}. 
This law is however much broader than the standards,
because a person can be hurt by colliding with an object that
indirectly harms people, or figuratively if material damages 
resulting from a crash are high. The same issue exists in European law: Loss can
be both economic and non-economic; it includes loss of income or
profit, burdens incurred and a reduction in the value of property; and
also physical pain and suffering and impairment of the quality of life~\cite{santosuosso_robots_2012}. Overall this means that safety should be
addressed for the entire operation of the robot.

%\begin{figure}
%\centering
% \fbox{\includegraphics[width=0.475\textwidth]{./img/StandardOverview.jpg}}
%\caption{A small overview of standards related to autonomous field robots.}
%\label{fig:StandardOverview}
%\end{figure}

Standards such as ISO 25119 for agriculture~\cite{iso25119_tractors_2010}, and ISO 13482, for mobile robots~\cite{iso13482_robots_2014}, are important for the overall
functional safety of the system. Additionally IEC 61496 is important for the
specific sensors~\cite{iec61496_safety_2013} and lastly ISO/DIS 18497
is an upcoming performance standard within agriculture to quantify
detection performances~\cite{iso/dis18497_agricultural_2014}. There
are many requirements to a computer vision system: it has to be able to observe a large area;
it must be fast, reliable and robust; and it is constrained to
function with low computing resources because it normally has to run on embedded
hardware, and might furthermore have lower priority than control and must not jam~\cite{aymeric_de_cabrol_concept_2008}.
%When looking at computer vision, Yang et al. \cite{yang_human_2012} achieves an Root-Mean-Square error of less than half a meter for detecting the distance to a human being. But an obstacle is not only a protruding element such as a human, obstacles has to be understood in its more broader definition, when looking at operational safety for a robot. The meaning of the word obstacle is according to [15] either "\textit{a situation or, an event etc. that makes it difficult for you to do or achieve something}" or "\textit{an object that is in your way and that makes it difficult for you to move forward}" \cite{hornby_oxford_2005}.This could refer to untraversable ground, which is a problem also for tractors today \cite{agricultural_statistics_board_2011_2013}, as well as obstacles like humans, animals and trees as examples. Because of safety concerns relating to autonomous robots, it is important to address these issues with standards. 
Standards within software for field robots are used to a very limited extent and not existing within computer vision \cite{ingibergsson_mmmi_2015}. %\TODO{Technical paper or not?}. 
%The functional-safety standards introduces safe states which are important during faults or malfunctions. Sensor in general has a risk of a failure. These failures can happen in all the sensors in the field robot. Most common sensors that are used are the encoders, gyros, sonars and cameras. Where faults can be sensor bias, locked in place and loss of calibration which are common sensor faults \cite{daigle_distributed_2007}.
This lack of standard is a paradox since autonomous mobile robots rely on robust
sensing to react, without robustness the robot may ``hallucinate'' and
respond inappropriately~\cite{murphy_handling_1999}. This issue puts
constraints not only on the software but also on the hardware. As an
example a RAW image has different degrees of being ``RAW''~\cite{brown_understanding_2015}. This difference in RAW can be seen
in different A/D converters, gains, and hardware image
optimizations. Because of this wide range input changes in hardware can be problematic, it
is therefore important for functional safety to look at software
safety verifications of the pipeline, and to give assurance about the
hardware and thereby verifying inputs and outputs.

MDE in robotics is an area that receives significant attention, such as
research within control~\cite{adam_towards_2014}, vision~\cite{hochgeschwender_declarative_2014} and general robot model-driven
development~\cite{hochgeschwender_model-based_2013,
  steck_model-driven_2011}. All the before-mentioned MDE methods
describe safety issues, however these issues are not addressed according
to any standards. Instead focus has been on quality, as with Reichardt
et. al who as an example deter from using code generation to improve
transparency~\cite{reichardt_software_2013}. Nevertheless we see code generation as
an improvement to reliability and the possibility of lowering the
demands for achieving certification, as is done by Bensalem
et. al.~\cite{bensalem_verifiable_2010}. Bensalem et. al. guarantee
safety using code generation, it is however not done according to any
standard. There exists several attempts to extend well-known MDE
environments such as RoboML and SysML to incorporate Failure Tree
Analysis (FTA) and safety analysis according to IEC 61508~\cite{yakymets_model-driven_2013}.
%Standards like ISO 13482 has an overview of hazards which needs to be addressed for mobile robots, an extension to this based on the material for ISO 13482 can be found Dogramdzi et. al. \cite{dogramadzi_environmental_2014}, where there exists descriptions of how to create the PHA and HRA for a robot system. 

\begin{figure*}[!t]
%jumps a page down before getting displayed.
\begin{center}
\centering
  \fbox{\includegraphics[width=0.8\textwidth]{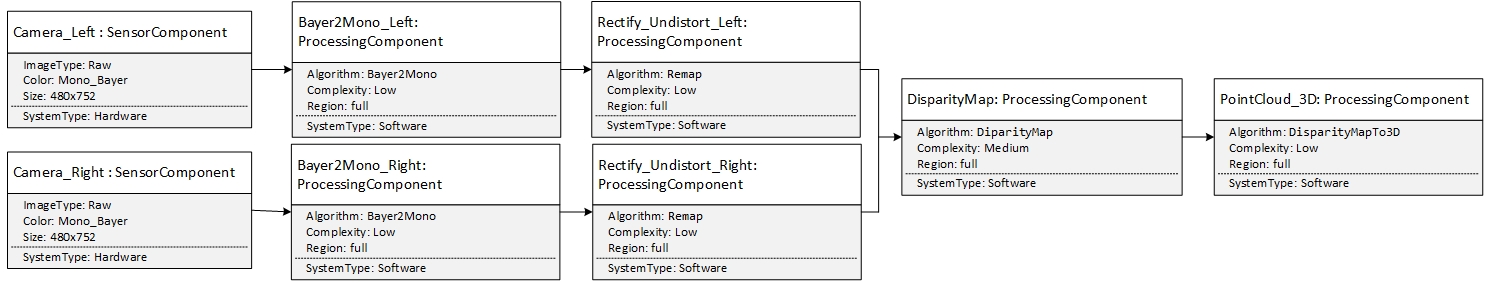}}

  \caption{Approximated object diagram of the software pipeline following the structure
  of RPSL.}\vspace*{-6mm} %The diagram is an approximation, e.g. not using the port object  from RPSL}
  \label{fig:VisionSWPipeline}
 
%  \label{fig:PublicationFrequencyAllPapersPerRTF}

\end{center}
\end{figure*}

\section{Problem Definition}
\label{sec:ProblemDefinition}

%UPS not critical, quick way to save significant space
%In this section, we present the concept and ideas for the study. We
%provide an overview of the system schematic. Finally we address
%functional safety in relation to a stereo vision camera, by discussing
%concepts for achieving functional safety through the system. We focus
%on software aspects in a vision pipeline shown in
%Fig. \ref{fig:VisionSWPipeline}. We will propose concepts that can be
%used to ascertain the working condition of the stereo camera system,
%in Fig. \ref{fig:VisionSWPipeline}.

\subsection{Safety Requirements}

To achieve any safety level a hazard analysis needs to be created,
often called a preliminary hazard analysis (PHA). At the stage of
creating a PHA the system configurations is unknown, during the
development the PHA becomes a hazard and risk analysis (HRA), when the
hazards are evaluate according to the standards functional safety
index, e.g., Agricultural Performance Level (AgPL) or System Integrity
Level (SIL).
%, this is done in relation to different environmental circumstances, e.g. machine moving or not and rain or fog etc.. This is the basis of the risk assessment which is done for each function in the system. 
Then, depending on the standard, either an iteration over the risk
assessment is done when the system has been redesigned, or the
assessment becomes the basis of the system design and puts constraints on
the system, e.g., redundancy as a hardware requirement.

Based on the introduction of field robots in an agricultural setting,
we propose to evaluate field robots within ISO 25119, for agriculture and
forestry~\cite{iso25119_tractors_2010}. For the evaluation of hazards
we refer to ISO 13482, which is for personal robots, but also covers
``multiple passengers'' or ``non-standing passengers'' or ``outdoor'' or
``uneven surfaces'' or ``not slow'' or ``not lightweight'' or ``autonomous''
(ISO 13482, Sect.~6.1.2.3, Person Carrier Robots, Type 3.2), %UPS removed [22]
which covers field robots. Annex A in ISO 13482~\cite{iso13482_robots_2014} gives an overview of hazards to be
evaluated. We refer to the broader system functionality from ISO 13482
specifically for type 3.2 field robot, Table~\ref{tab:ISO13482PerformanceLevels}.

\begin{table}[!t]
\centering
\caption{Performance levels for field robot.}
\label{tab:ISO13482PerformanceLevels}
\begin{tabular}{|l|c|}
\hline
\textbf{Safety functions of robots} & \textbf{Type 3.2}\\
\hline \hline
Emergency Stop & d\\ \hline
Protective Stop & e\\ \hline
Limits to workspace(incl. forbidden area avoidance) & e\\ \hline
safety-related speed control & e\\ \hline
safety-related force control & N/A\\ \hline
Hazardous collision avoidance & e\\ \hline
Stability Control (incl. overload protection) & d\\
\hline
\end{tabular}\vspace*{-6mm}
\end{table}

Note that if the field care robot is inherently unstable, \emph{PL e} is required.
Moreover, the control system shall achieve \emph{PL e}, but this might not be
achievable for sensing mechanisms. In this case, the risks caused by
systematic failure of sensors shall be reduced as low as reasonably
practicable.  
When assessing functional safety, then the entire path for the
function needs to be the same level, e.g., sensor input to control
output. This means that the vision system, based on Table~\ref{tab:ISO13482PerformanceLevels}, would reach an e-level. This has
implications both on hardware redundancy and software development
practices; we assume that the hardware system can be acquired.  

We hypothesize that since the requirements are on functional safety
and not on performance, then the algorithms are not directly
implicated by the safety level. Moreover, if the detection
algorithms perform as proposed in ISO 18497, and are developed
according to Misra~\cite{misra_misra-c_2012} software development
practices, then it would be acceptable. The functional safety
requirements would therefore be addressed by developing a system that
can test and verify the functionality of the functions, and thereby
ensure that the algorithms and sensor are functioning. This
interpretation is also in-line with the standard IEC 61496, inferred
from the descriptions of the different types of Electro-Sensitive
Protective Equipment (EPSE). The EPSE types addresses monitoring of
the system working condition, and ability to perform, e.g. response
time. Another point is based on the \emph{Note} for 
Table~\ref{tab:ISO13482PerformanceLevels}, considering the sensors, it might
be not achievable to get all parts of the sensor certified, in this
case the highest level should be achieved. The essence of our
interpretation of implementing functional safety is that one must
ensure that all the preconditions for an action hold before that
action can be safely and correctly executed, similar to the notion put
forth by Rahimi et. al.~\cite{mansour_rahimi_framework_1991}. 

We draw the following %important 
aspects on EPSE from
IEC 61496: Detection Zone, Detection capability, adjustment (failure to danger
not possible). These points needs to be addressed in order to
ascertain the functionality of the sensor. Specifically for software
it is stated that it should be developed in accordance with IEC
61508-3 or ISO 13849. IEC 61508 covers all machines, however forestry
and agricultural machinery has ISO 25119 which is a type C standard,
i.e., shall be used within the covered machinery types. Therefore
software for our system should be developed according to ISO 25119,
supporting our prior commitment to this standard. %\\

%In relation to software ISO 25119 covers the whole cycle from \emph{software architecture and design} to \emph{software integration and testing} and finally \emph{software safety validation}. These subjects covers the use of Commercial of the shelf software (COTS), choosing programming language, design methods, coding standards and inspection. For testing there are requirements on using dynamical, statical, functional, performance and interface testing. All these subjects depend on the overall configuration of the system, hardware components, configuration and diagnostic coverage and finally \emph{software requirement level}.

\section{Proposal and Experiments}
\label{sec:ExpResult}

%We will start with a discussion and give proposals on solutions for achieving functional safety within the vision pipeline, ending with the implementation of some of the functionality in a test setup to verify the conceptual ideas.

\subsection{Proposal}
\label{sec:AnalysisProcedures}

Based on the discussed safety and the scenario overview, we %will now
propose a systematic MDE-based approach to
%focus on how to introduce 
introducing safety in a vision pipeline. As an example, we use a simple
vision pipeline that generates a 3D point cloud from two RAW
images. The pipeline uses a debayer filter to convert the RAW images into grayscale images,
they are then rectified and undistorted using camera calibration information, to
enable the creation of disparity maps. Finally the disparity maps
are converted to a 3D point cloud. The pipeline is described as a UML
model using metamodels from RPSL~\cite{hochgeschwender_declarative_2014},
the model is shown in Fig.~\ref{fig:VisionSWPipeline}. We would in this pipeline
hypothesize that it is possible to achieve a higher safety level by
using an extended version of RPSL that has the ability to annotate the model with safety and validity
requirements.  These annotations would be used to generate code that continuously checks the integrity and correctness of the vision pipeline.  A camera is a sensor influenced by many
factors, lenses, processors and software, so the safety functions should
therefore not only address software errors but also both mechanical and
hardware errors.
%Dogramdzi et. al. has an example where a camera hazard is that there is no input\cite{dogramadzi_environmental_2014}, the issue is that the input from a camera can be wrong and still give an input, which means that a camera can have multiple errors. The errors can range from an issue with lenses, to computation of an output. 
%For example, there are many things that could cause the video signal from a camera to be interrupted, but distinguishing between them is unnecessary, because the only way to recover from any of them is to give up on that camera and try to use another sensor. \cite{murphy_handling_1999}

%Therefore
Concretely, we propose two simple methods of improving knowledge about
faulty cases and verifying the functionality of the system, by introducing
%the following two ideas:
declarative rules that specify the behavior of the pipeline according to:
%
%\begin{itemize}
%\item
histograms of grayscale images and
%\item
known landmark recognition.
%\end{itemize} 
%
The rules should be introduced in a simple way by using a
domain-specific language (DSL). A DSL would give flexibility to the
user, meanwhile using code generation it would provide high
reliability in the safety functions. The two methods introduced above
would contribute to knowledge about the working condition of the
software pipeline and the verification of the input and the working
condition of the camera.

The histogram analysis could give a description of how much of
the intensity spectrum is used and how large a spread exists between the different intensity values. This information would supply basic information about the sensor's
working condition and if the lens is covered (in that case most pixels would have a very low intensity). The proposed rule DSL could 
express the histogram concepts as follows: %with the following simple rules:

%\begin{figure*}
%\hrulefill
\begin{small}
\begin{verbatim}
h=Bayer2Mono_Left.output.histogram;
length(nonempty(h.bins))/length(h.bins)>0.1;
max(h)-min(h)>1000p;
\end{verbatim}
\end{small}
%Histogram:
%rule length(Camera_Left.Output.histogram.bins.Filled)>10%;
%rule (Camera_Left.Output.histogram.Max-Camera_Left.Output.histogram.Min)>1000p;
%\hrulefill
%\end{figure*}

\noindent
%TODO: describe what the rules are specifying
The rule extracts a greyscale histogram of the specific cameras image, i.e. 
Camera\_Left (more precisely the bayer filtering result based on an image from Camera\_Left) . The found histogram consists of bins representing the number 
of pixels for each tonal value. The division gives the ratio of bins 
with at least one pixel. The rule imposes that 10 percent of the bins 
should be filled to make the image trustworthy.
This rule could also be used directly on the RAW image from ``Camera\_Left'' filtering in 
Fig.~\ref{fig:VisionSWPipeline}. This would allow to separate the verification process between camera problems and processing errors. Since the RAW image approach would require a more complicated technical description we stayed with the grayscale case here since we believe that this gives the reader a better understanding.

The second idea is to introduce a known landmark in a defined area of
the real world that the camera observes. This known landmark would then result in some 3D points in
a defined area. If the points are not found in that region, then 
there is an error in the vision pipeline, and the results from the system
cannot be trusted. Again the proposed rule DSL could introduce this check by 
the following rules:

%\begin{figure*}
%\hrulefill
%//Landmark positioning: UPS: can't understand?
%Camera_Left.Landmark.pos.rect(0,Camera_Left.height,100,Camera_Left.height-100);
%//Landmark:
\begin{small}
\begin{verbatim}
length(PointCloud_3D.output.
  inArea(Camera_Left_Landmark))>900;
\end{verbatim}
\end{small}
%Landmark positioning:
%rule Camera_Left.Landmark.pos.rect(0,Camera_Left.height,100,Camera_Left.height-100);
%Landmark:
%rule PointCloud_3D.pointsFound(p => p >= 10).inArea(CameraLeft.Landmark);
%\hrulefill
%\end{figure*}
The rule uses the resulting point cloud of the vision pipeline, Figure 
\ref{fig:VisionSWPipeline}. The rule uses the output and extracts the points 
from a specific area, in this case Camera\_Left\_Landmark. The rule then 
specifies that at least 901 3D points should be found in this area. Since it 
is known that the landmark should exist in the specified area the rule would 
verify the functionality of the functions and focus of the lens. 

\subsection{Experimental Setup and Results}

Our experimental setup consists of a camera with CAN and USB interface. The 
camera is a CLAAS Cam Pilot stereo camera~\cite{claas_culti_2015}. %see Figure
%\ref{fig:ClaasCultiCam}. 
The interface to the camera uses
CLAAS hardware to convert USB signal to messages for controlling
the camera. The raw pictures are extracted using the USB interface to
the camera and enables easy analysis on a PC.

%% \begin{figure}
%% \centering
%%  \fbox{\includegraphics[scale=0.2]{./img/ClaasCultiCam.jpg}}
%% \caption{Stereo camera supplied by CLAAS for this implementation.}
%% \label{fig:ClaasCultiCam}
%% \end{figure}

%\subsection{Results}

\begin{figure}
\centering
\begin{subfigure}[b]{0.2\textwidth}

    \includegraphics[width=\textwidth]{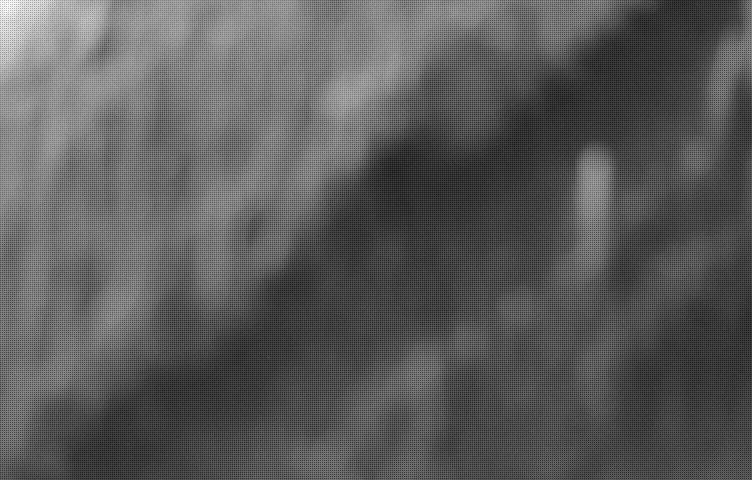}
    \caption{Left lens, covered.}
    
  \end{subfigure}
  \begin{subfigure}[b]{0.2\textwidth}

    \includegraphics[width=\textwidth]{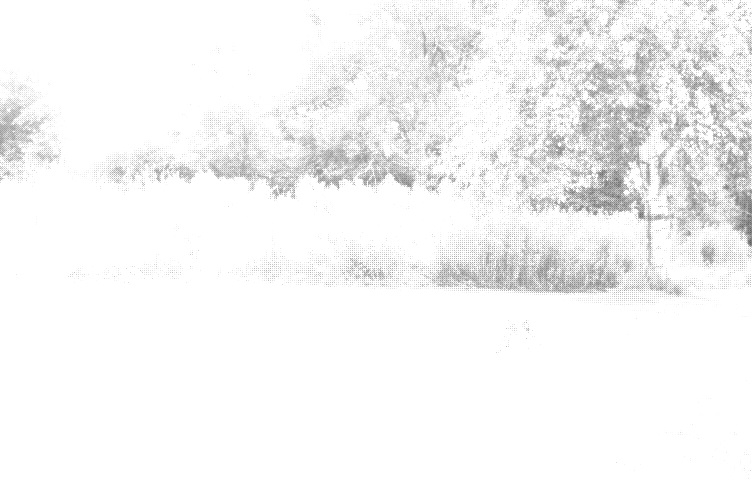}
    \caption{right lens, overexposed.}

  \end{subfigure}
  \begin{subfigure}[b]{0.2\textwidth}

    \includegraphics[width=\textwidth]{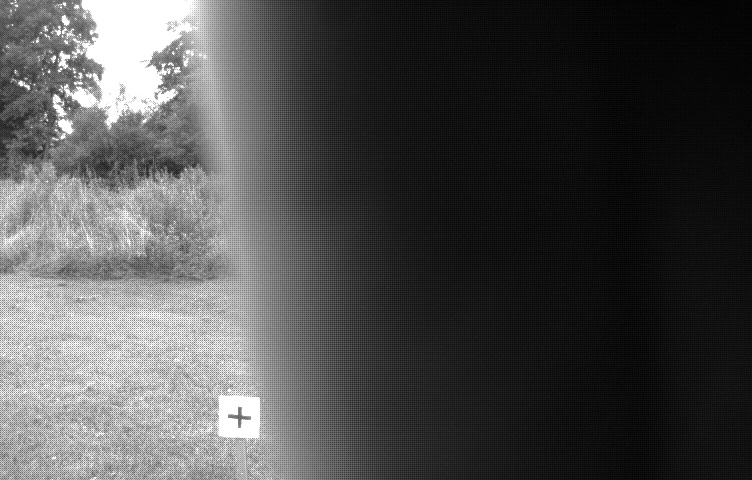}
    \caption{Left lens, partial cover}
    
  \end{subfigure}
  \begin{subfigure}[b]{0.2\textwidth}

    \includegraphics[width=\textwidth]{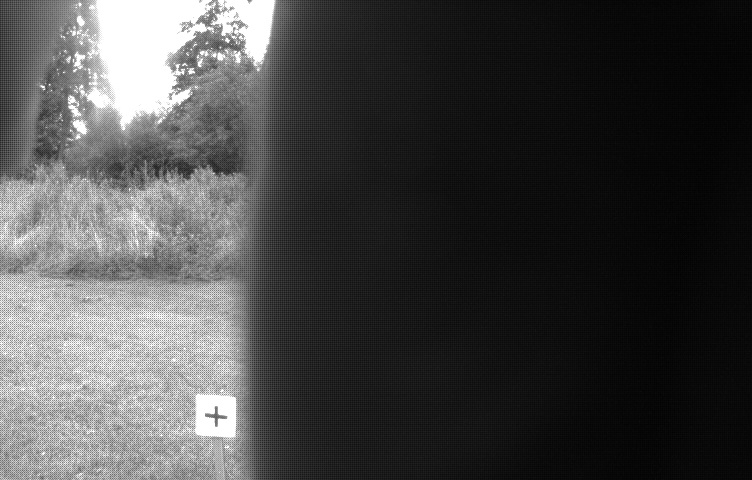}
    \caption{right lens, partial cover}

  \end{subfigure}
  \caption{Sample images taken with Cam Pilot camera with introduction of 
  different faults.}\vspace*{-6mm}
  \label{fig:ImagesofScenes}
\end{figure}

The above rules were implemented by hand in the vision pipeline
(Fig. \ref{fig:VisionSWPipeline}). The simple function of analysing
histograms of the raw pictures has made it possible to ascertain that
the input falls within expected ranges. Which made it possible to catch 
cases such as (a) and (b) shown on Figure \ref{fig:ImagesofScenes}. As an 
example, the software is able to prompt
an error message, and will not let the system continue with the analysis 
of the images. This error prompt could be used in field robots in the
decision module to stop the robot from moving, and signalling an
operator to come and solve the issue. The landmark that was
introduced was a white square surface with a black cross in the lower part 
of the field of view of the camera. This
enabled a test of the entire software pipeline, by evaluating the
output of the 3D point cloud to have a result in the expected
area. If the landmark is not detected the system will give an error of the 
lens and software pipeline, giving a higher reliability. This was quite 
robust, nevertheless it was possible to create scenarios where the results 
were trusted wrongly, as with the case for the pictures (c) and (d) on 
Figure \ref{fig:ImagesofScenes}. In this case neither of the rules caught 
the fault. We however believe that extending the histogram to look at 
distributions may have a change of catching issues of partially covered 
lenses.
The results verify that the functionality of the camera can be improved 
with simple rules. However extensions to the rules are needed, such as the proposal
for looking at distributions of exposures, perhaps extended to the entire colour range.

%the overexposures and underexposures. %Which 
%in turn could be extended to the entire colour range. 
%\begin{figure}[!t]
%\centering
%\subfigure[caption]{\includegraphics[width = 2in]{./img/00477-RawLeft.jpg}}
%\subfigure[caption]{\includegraphics[width = 2in]{./img/00477-RawRight.jpg}}
%%\begin{subfigure}{.5\textwidth}
%%  \centering
%%  \includegraphics[width=.4\linewidth]{./img/00477-RawLeft.jpg}
%%  \caption{A subfigure}
%%  \label{fig:sub1}
%%\end{subfigure}%
%%\begin{subfigure}{.5\textwidth}
%%  \centering
%%  \includegraphics[width=.4\linewidth]{./img/00477-RawRight.jpg}
%%  \caption{A subfigure}
%%  \label{fig:sub2}
%%\end{subfigure}
%%\caption{A figure with two subfigures}
%%\label{fig:test}
%\end{figure}

\section{Conclusion}
\label{sec:Conc}

This paper has introduced concepts and ideas of how to improve safety
in relation to standards. The idea was based on the notion of
splitting the performance and functionality requirements of the system
and thereby minimizing certification requirements of a sensor system.
Our implementation and tests gives an indication that our approach to
annotate safety requirements in a vision pipeline leads to improved
safety.
%
%\subsection{Future work}
%
In terms of future work, this is
%the presented methods are 
a first step toward introducing safety
certification within vision systems for field robots. To further investigate
the concept and validity of the approach, %it would be interesting to
%have a discussion with certification authorities.
we are currently discussing it with certification authorities to understand 
the extend of the safety needed.
Our safety concept could be extended by introducing an ``imaginary
cage''. The idea is to make the system trustworthy enough that the
cameras could make a perimeter around the robot, which for example
could be calculated on the basis of the approach suggested by Täubig
et. al.~\cite{taubig_guaranteeing_2012}.

%% This safety concept could be extended by introducing an ``imaginary
%% cage''. The idea is to make the system trustworthy enough that the
%% cameras could make a perimeter around the robot. This perimeter could
%% be calculated on the basis of Täubig et. al.'s idea of Guaranteeing
%% functional safety \cite{taubig_guaranteeing_2012}. The speed and
%% detection zones could further be extended by the notion of
%% \emph{passive motion safety} \cite{bouraine_provably_2012} in
%% connection with Guaranteeing a safety zone
%% \cite{taubig_guaranteeing_2012}. Introducing these capabilities within
%% the verification system for the sensor, then the system could achieve
%% the functional safety level of the different stop types, categorised
%% in Table \ref{tab:ISO13482PerformanceLevels}.  Returning to the safety
%% concepts introduced with histograms and landmarks. This could be
%% extended with 2D-FFT, to give an indication of the sharpness of an
%% images. An idea could also be to calculate the pixel noise or image
%% optical flow to ascertain changes and working condition of the
%% sensor. Another approach to verify the functionality of individual
%% components/algorithms could be to use stored images and their results,
%% to see if the output matches earlier recorded results. It might also
%% be interesting to combine information from different sensors, such as
%% IMU, radar or lidar, to improve reliability and introduce new safety
%% concepts.

\nocite{mistry_towards_2014}
\nocite{malm_safety-critical_2011}
\bibliographystyle{ieeetr}
%\bibliography{ZotOutput}{}

\begin{thebibliography}{10}

\bibitem{patrick_j._griffin_safety_2013}
{Patrick J. Griffin}, ``Safety and {Health} in {Agriculture} "{Farming} - a
  hazardous occupation – how to improve health \& safety?",'' 2013.

\bibitem{agricultural_statistics_board_agricultural_2013}
{Agricultural Statistics Board}, ``Agricultural {Safety}: 2009 {Injuries} to
  {Adults} on {Farms},'' 2013.

\bibitem{yang_remote_2008}
S.-Y. Yang, S.-M. Jin, and S.-K. Kwon, ``Remote control system of industrial
  field robot,'' in {\em {IEEE} {Int.} {Conf.} on {Industrial}
  {Informatics}}, pp.~442--447, IEEE, 2008.

\bibitem{ingibergsson_mmmi_2015}
J.~T.~M. Ingibergsson, U.~P. Schultz, and M.~Kuhrmann, ``On the use of safety
  certification practices in autonomous field robot software development: A
  systematic mapping study,'' Tech. Rep. MMMI TR-2015-1, 2015.
\newblock Submitted for publication.

\bibitem{carlson_follow-up_2004}
J.~Carlson, R.~R. Murphy, and A.~Nelson, ``Follow-up analysis of mobile robot
  failures,'' in {\em {ICRA}}, vol.~5, pp.~4987--4994, IEEE, 2004.

\bibitem{reichardt_software_2013}
M.~Reichardt, T.~Föhst, and K.~Berns, ``On software quality-motivated design
  of a real-time framework for complex robot control systems,'' in {\em
  Int. {Workshop} on {Software} {Quality} and {Maintainability}},
  2013.

\bibitem{schmidt_guest_2006}
D.~Schmidt, ``Guest {Editor}'s {Introduction}: {Model}-{Driven}
  {Engineering},'' {\em Computer}, vol.~39, pp.~25--31, Feb. 2006.

\bibitem{steck_model-driven_2011}
A.~Steck, A.~Lotz, and C.~Schlegel, ``Model-driven engineering and run-time
  model-usage in service robotics,'' in {\em {Proc.} of {ACM GPCE}},
  pp.~73--82, ACM, 2011.

\bibitem{hochgeschwender_declarative_2014}
N.~Hochgeschwender, S.~Schneider, H.~Voos, and G.~K. Kraetzschmar,
  ``Declarative {Specification} of {Robot} {Perception} {Architectures},'' in
  {\em SIMPAR},
  pp.~291--302, Springer, 2014.

\bibitem{aguilera_visual_2006}
J.~Aguilera, D.~Thirde, M.~Kampel, M.~Borg, G.~Fernandez, and J.~Ferryman,
  ``Visual surveillance for airport monitoring applications,'' in {\em
  Proc. of the 11th {Computer} {Vision} {Winter} {Workshop}}, pp.~6--8,
  2006.

\bibitem{cheng_state---art_2011}
P.~H. Cheng, ``The {State}-of-the-{Art} in the {USA},'' in {\em Autonomous
  {Intelligent} {Vehicles}}, Advances in {Computer} {Vision} and {Pattern}
  {Recognition}, pp.~13--22, Springer London, Jan. 2011.

\bibitem{aymeric_de_cabrol_concept_2008}
{Aymeric De Cabrol}, {Thibault Garcia}, {Patrick Bonnin}, and {Maryline
  Chetto}, ``A concept of dynamically reconfigurable real-time vision system
  for autonomous mobile robotics,'' {\em Int. Journal of Automation
  and Computing}, vol.~5, 2008.

\bibitem{yang_human_2012}
L.~Yang and N.~Noguchi, ``Human detection for a robot tractor using
  omni-directional stereo vision,'' {\em Computers and Electronics in
  Agriculture}, vol.~89, pp.~116--125, 2012.

\bibitem{santosuosso_robots_2012}
A.~Santosuosso, C.~Boscarato, F.~Caroleo, R.~Labruto, and C.~Leroux, ``Robots,
  market and civil liability: {A} {European} perspective,'' in {\em {RO}-{MAN},
  2012 {IEEE}}, pp.~1051--1058, IEEE, 2012.

\bibitem{iso26262_road_2011}
{ISO26262}, {\em Road {Vehicles} {Functional} {Safety}}.
\newblock ISO, 2011.

\bibitem{iso25119_tractors_2010}
{ISO25119}, ``Tractors and machinery for agriculture and forestry –
  {Safety}-related parts of control systems,'' ISO
  25119-2010.

\bibitem{hedenberg_safety_2011}
K.~Hedenberg and B.~Åstrand, ``Safety standard for mobile robots - a proposal
  for 3d sensors,'' (Sweden), 2011.

\bibitem{en1525_safety_1998}
{EN1525}, {\em Safety of industrial trucks. {Driverless} trucks and their
  systems}.
\newblock EN, 1998.

%\bibitem{asimov_i_1950}
%I.~Asimov, {\em I, {Robot}}.
%\newblock Garden City, N.Y.: Doubleday, 1950.

\bibitem{iso13482_robots_2014}
{ISO13482}, {\em Robots and robotic devices - {Safety} requirements for
  personal care robots}.
\newblock ISO, 2014.

\bibitem{iec61496_safety_2013}
{IEC61496}, {\em Safety of machinery - {Electro}-sensitive protective
  equipment}.
\newblock IEC, 2013.

\bibitem{iso/dis18497_agricultural_2014}
{ISO/DIS18497}, {\em Agricultural machinery and tractors – {Safety} of highly
  automated machinery}.
\newblock ISO, 2014.

\bibitem{murphy_handling_1999}
R.~R. Murphy and D.~Hershberger, ``Handling sensing failures in autonomous
  mobile robots,'' {\em The Int. J. of Robotics Research},
  vol.~18, no.~4, pp.~382--400, 1999.

\bibitem{brown_understanding_2015}
M.~S. Brown and S.~J. Kim, ``Understanding the {In}-{Camera} {Image}
  {Processing} {Pipeline} for {Computer} {Vision},'' 2015.

\bibitem{adam_towards_2014}
S.~Adam, M.~Larsen, K.~Jensen, and U.~P. Schultz, ``Towards {Rule}-{Based}
  {Dynamic} {Safety} {Monitoring} for {Mobile} {Robots},'' in {\em SIMPAR}, Vol.~8810 in
  {LNCS}, pp.~207--218, Springer, 2014.

\bibitem{hochgeschwender_model-based_2013}
N.~Hochgeschwender, L.~Gherardi, A.~Shakhirmardanov, G.~K. Kraetzschmar,
  D.~Brugali, and H.~Bruyninckx, ``A model-based approach to software
  deployment in robotics,'' in {\em IROS}, 2013 {IEEE}/{RSJ}, pp.~3907--3914,
  2013.

\bibitem{bensalem_verifiable_2010}
S.~Bensalem, L.~da~Silva, M.~Gallien, F.~Ingrand, and R.~Yan, ``Verifiable and
  correct-by-construction controller for robots in human environments,'' in
  {\em {DRHE} 2010 dependable robots in human environments, seventh {IARP}
  workshop on technical challenges for dependable robots in human
  environments}, 2010.

\bibitem{yakymets_model-driven_2013}
N.~Yakymets, S.~Dhouib, H.~Jaber, and A.~Lanusse, ``Model-driven safety
  assessment of robotic systems,'' in {\em IROS}, 2013 {IEEE}/{RSJ} , pp.~1137--1142,
  2013.

\bibitem{misra_misra-c_2012}
{MISRA}, {\em {MISRA}-{C} {Guidelines} for the {Use} of the {C} {Language} in
  {Critical} {Systems}}.
\newblock Motor Industry Software Reliability Assoc., 2012.

\bibitem{mansour_rahimi_framework_1991}
{Mansour Rahimi} and {Xia Xiadong}, ``A framework for software safety
  verification of industrial robot operations,'' {\em Computers \& Industrial
  Engineering}, vol.~20, no.~2, pp.~279--287, 1991.

\bibitem{claas_culti_2015}
{CLAAS}, ``Culti {Cam} {Picture},'' 2015.

\bibitem{taubig_guaranteeing_2012}
H.~Täubig, U.~Frese, C.~Hertzberg, C.~Lüth, S.~Mohr, E.~Vorobev, and
  D.~Walter, ``Guaranteeing functional safety: design for provability and
  computer-aided verification,'' {\em Autonomous Robots}, vol.~32,
  pp.~303--331, Apr. 2012.

\bibitem{mistry_towards_2014}
A.~Winfield, C.~Blum, and W.~Liu, ``Towards an {Ethical} {Robot}: {Internal}
  {Models}, {Consequences} and {Ethical} {Action} {Selection},'' in {\em
  Advances in {Autonomous} {Robotics} {Systems}} (M.~Mistry, A.~Leonardis,
  M.~Witkowski, and C.~Melhuish, eds.), vol.~8717 of {\em Lecture {Notes} in
  {Computer} {Science}}, pp.~85--96, Springer International Publishing, 2014.

\bibitem{malm_safety-critical_2011}
T.~Malm, M.~Vuori, J.~Rauhamäki, T.~Vepsäläinen, J.~Koskinen, J.~Seppälä,
  H.~Virtanen, and M.~Hietikko, {\em Safety-critical software in machinery
  applications}.
\newblock VTT, 2011.

\end{thebibliography}

\end{document}